\documentclass[letterpaper]{article} 
\usepackage[]{aaai24}  
\usepackage{times}  
\usepackage{helvet}  
\usepackage{courier}  
\usepackage[hyphens]{url}  
\usepackage{graphicx} 
\usepackage{natbib}  
\usepackage{caption} 
\usepackage{float, subcaption}
\usepackage{graphicx}
\usepackage[ruled,vlined]{algorithm2e}
\usepackage[tableposition=top]{caption}
\usepackage{booktabs}
\usepackage{multirow}
\usepackage{amsmath}
\usepackage{array}

\usepackage{adjustbox}
\usepackage[table]{xcolor}
\usepackage{soul}
\sethlcolor{gray!60}
\usepackage{comment}
\usepackage[inline]{enumitem}

\usepackage{pifont}

\newcolumntype{L}[1]{>{\raggedright\let\newline\\\arraybackslash}m{#1}}
\newcolumntype{C}[1]{>{\centering\let\newline\\\arraybackslash}m{#1}}
\newcolumntype{R}[1]{>{\raggedleft\let\newline\\\arraybackslash}m{#1}}

\usepackage[none]{hyphenat}
\usepackage{newfloat}
\usepackage{listings}
\DeclareCaptionStyle{ruled}{labelfont=normalfont,labelsep=colon,strut=off} 
\floatstyle{ruled}
\newfloat{listing}{tb}{lst}{}
\floatname{listing}{Listing}
\title{WinTSR: A Windowed Temporal Saliency Rescaling Method for Interpreting \\Time Series Deep Learning Models}
\author {
    Md. Khairul Islam\textsuperscript{\rm 1},
    Judy Fox\textsuperscript{\rm 1, 2}
}
\affiliations {
    \textsuperscript{\rm 1}Computer Science Dept, University of Virginia\\
    \textsuperscript{\rm 2}School of Data Science, University of Virginia\\
    mi3se@virginia.edu, ckw9mp@virginia.edu
}

\usepackage{bibentry}

\begin{document}

\maketitle

\begin{abstract}

Interpreting complex time series forecasting models is challenging due to the temporal dependencies between time steps and the dynamic relevance of input features over time. Existing interpretation methods are limited by focusing mostly on classification tasks, evaluating using custom baseline models instead of the latest time series models, using simple synthetic datasets, and requiring training another model. We introduce a novel interpretation method, \textit{Windowed Temporal Saliency Rescaling (WinTSR)} addressing these limitations. WinTSR explicitly captures temporal dependencies among the past time steps and efficiently scales the feature importance with this time importance. We benchmark WinTSR against 10 recent interpretation techniques with 5 state-of-the-art deep-learning models of different architectures, including a time series foundation model. We use 3 real-world datasets for both time-series classification and regression. Our comprehensive analysis shows that WinTSR significantly outperforms other local interpretation methods in overall performance. Finally, we provide a novel, open-source framework to interpret the latest time series transformers and foundation models. 

\end{abstract}

\section{Introduction}

Time-series deep learning models have achieved unprecedented performance in recent years. However, the lack of explainability remains a key challenge for their widespread use. Explanations provide the necessary transparency to make reliable decisions, especially in sensitive data such as healthcare, finance, energy, traffic, and many other domains \citep{benidis2022deep}. Explanations can be \textit{Global}, the logic and reasoning of the entire model, or \textit{Local}, for a specific instance. \textit{Post-hoc} interpretation methods are generally applied after the model has already been trained, while \textit{In-hoc} methods work during the model training time. \textit{Model-agnostic} methods work on black-box models and do not require specific model architecture to work. Our proposed interpretation method is local, post-hoc, and model-agnostic.

Unlike image and text data, the interpretability of multivariate time series models is relatively under-explored and difficult to visualize. By explaining time series models, one can highlight the importance of input features to the prediction of the model \citep{rojat2021explainable}, find intuitive patterns \cite{lim2021temporal}, and visualize saliency maps \cite{leung2022temporal}. This work focuses on the local interpretation techniques for deeper insights into the importance of temporal features. Interpretation methods commonly estimate how relevant each input feature is to the model's output. However, existing works are limited by: (1) benchmarking using simple baseline models (e.g. LSTM, GRU), not recent SOTA time series models that are used in practice (2) focusing mostly on classification tasks only (3) not efficiently capturing temporal dependency (4) train another model to interpret one model. 

\begin{figure*}[!htb]
 \centering
\includegraphics[width=0.75\textwidth]{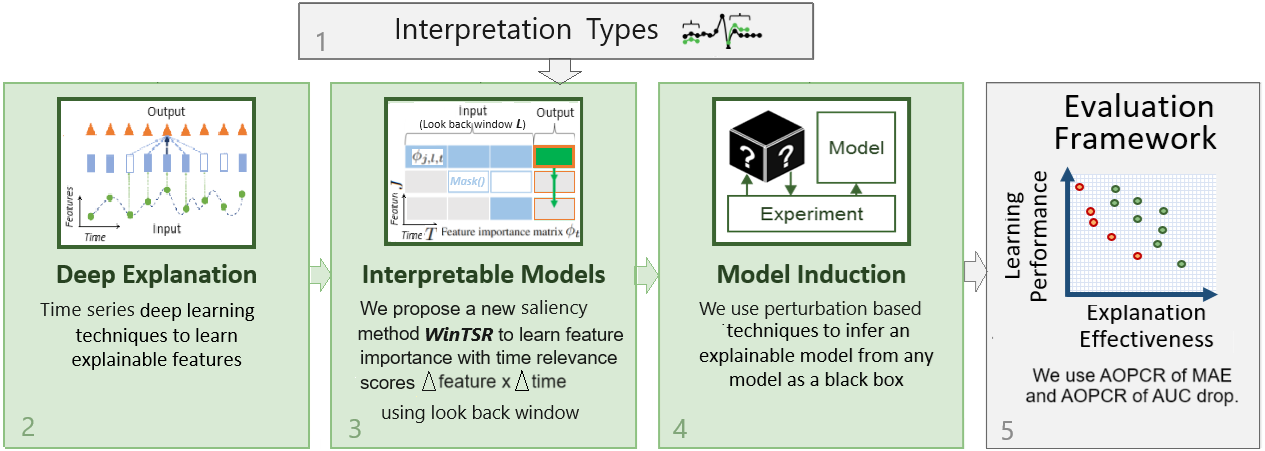}
\caption{An overview of our framework with the proposed  \textit{Windowed Temporal Saliency Rescaling} (\textbf{WinTSR}) method.}
\label{fig:architecture}
\end{figure*}
\begin{table*}[!hbt] 
    \small 
    \centering
    \caption{Summary of related works on local interpretation methods for time series. Most works focus on simple synthetic datasets for classification tasks with baseline models (e.g. GRU or LSTM). }
 \resizebox{.99\textwidth}{!}{   
    \begin{tabular}{ |l|C{.9cm}C{1.5cm}|C{3.5cm}|ccc|cc|} \hline
        \multirow{2}{*}{\textbf{Related Work}} &  \multicolumn{2}{c|}{\textbf{Method}} & \multirow{2}{*}{\textbf{Model}} & \multicolumn{3}{c|}{\textbf{Dataset}} &\multicolumn{2}{c|}{\textbf{Task}}   \\   \cline{2-3} \cline{5-9}
        &  {Gradient} & {Perturb.}  &
         & {Real} & {Synthetic} & Time series & {Class.} & {Regr.} 
        \\ \hline
        \textbf{FA} \cite{Suresh2017ClinicalIP}  &  & {\LARGE \textbullet}   & LSTM, CNN  & {\LARGE \textbullet} & &  & {\LARGE \textbullet} &   \\
        \textbf{AFO} \cite{Tonekaboni2020WhatWW}&  &  {\LARGE \textbullet}  &  RNN & {\LARGE \textbullet} & {\LARGE \textbullet} & & {\LARGE \textbullet} &    \\
        \textbf{FP} \cite{molnar2020interpretable} &  & {\LARGE \textbullet} &  SVM & {\LARGE \textbullet} & &  & {\LARGE \textbullet} & {\LARGE \textbullet}   \\
        \textbf{IG} \cite{Sundararajan2017AxiomaticAF} &   {\LARGE \textbullet} &    & ImageNet, LM, GCNN   & {\LARGE \textbullet} &  &  & {\LARGE \textbullet} &     \\
        \textbf{GS} \cite{lundberg2017unified}              &  {\LARGE \textbullet} &   & CNN  & {\LARGE \textbullet} & &  & {\LARGE \textbullet} &     \\
        \textbf{TSR} \cite{ismail2020benchmarking} &  {\LARGE \textbullet}     &  & LSTM, CNN, Transformer & & {\LARGE \textbullet} & {\LARGE \textbullet} & {\LARGE \textbullet} &    \\ 
        \textbf{DM} \cite{Crabbe2021ExplainingTS} &   &    {\LARGE \textbullet}  &  RNN & {\LARGE \textbullet} & {\LARGE \textbullet} & {\LARGE \textbullet} &{\LARGE \textbullet} & {\LARGE \textbullet}   \\ 
        \textbf{EM} \cite{enguehard2023learning} &    &   {\LARGE \textbullet}  &  RNN (1 layer GRU) & {\LARGE \textbullet} & {\LARGE \textbullet} & {\LARGE \textbullet} & {\LARGE \textbullet} &  \\ 
        \textbf{WinIT} \cite{leung2022temporal}  &    & {\LARGE \textbullet}  &  GRU, LSTM, ConvNet & {\LARGE \textbullet} &  {\LARGE \textbullet} &{\LARGE \textbullet} & {\LARGE \textbullet} &   \\
        \textbf{ContraLSP} \cite{liu2024explaining} &     &  {\LARGE \textbullet}  & RNN (1 layer GRU) & {\LARGE \textbullet} & {\LARGE \textbullet} & {\LARGE \textbullet} & {\LARGE \textbullet}
        &    \\ 
        \rowcolor{gray!35}       
        \textbf{WinTSR} (Ours)   &   & {\LARGE \textbullet}  &  DLinear, SegRNN, MICN, iTransformer, CALF & {\LARGE \textbullet} &  & {\LARGE \textbullet} &{\LARGE \textbullet} & {\LARGE \textbullet} \\ 
        \hline
    \end{tabular}
    }
\label{tab:TS-SaliencyInterpretation-benchmark}
\end{table*}

We propose the \textit{Windowed Temporal Saliency Rescaling (WinTSR)} method to address these challenges. It is benchmarked using the latest time series models of different architectures (including an LLM-based foundation model), tested on both classification and regression tasks, efficiently considers the temporal importance when interpreting, and does not require training a separate model. Our overall framework is summarized in Figure \ref{fig:architecture}. In short, our contributions are,

\begin{itemize}
   \item[1)] A novel local interpretation method named \textit{WinTSR} that calculates the time importance along the look-back window and rescales the feature importance based on it. This significantly improves the performance by capturing the delayed impact between input and output, while still being relatively fast.
   \item[2)] Extensive analysis with 3 real-world datasets for classification and regression tasks, and interpretation evaluation using robust metrics.
    \item[3)] Benchmark \textit{WinTSR} with 5 state-of-the-art time series models (DLinear, MICN, SegRNN, iTransformer) including a foundation model (CALF) to demonstrate that WinTSR is generalizable and consistently outperforms in different model architectures. 
    \item[4)] A unified open-source framework that includes 20+ recent time series models (including 3 foundation models, Appendix \ref{sec:available_models}) with 10+ popular interpretation methods. Also, visualize and compare the multivariate temporal trends interpreted by different methods. \textbf{Code available} as open source in the Github \footnote{https://github.com/khairulislam/Timeseries-Explained}.
\end{itemize}

\section{Related Works \label{sec:related_works}}

Time series interpretation methods cover a wide range of tasks and datasets \cite{rojat2021explainable, turbe2023evaluation}. Table \ref{tab:TS-SaliencyInterpretation-benchmark} summarizes the comparisons of our work with the related methods. \textit{Gradient based} methods, such as Integrated Gradients \citep{Sundararajan2017AxiomaticAF}, and GradientSHAP \citep{Erion2019LearningEM} use the gradient of the model predictions to input features to generate importance scores. \textit{Perturbation based} methods, such as Feature Ablation \citep{Suresh2017ClinicalIP}, and Augmented Feature Occlusion \cite{Tonekaboni2020WhatWW} replace a feature or a set of features from the input using some baselines or generated masks and measure the importance based on the model output change. These methods are mainly proposed for image or language models and \textbf{do not consider the temporal dependencies in the time series data}.

Recent works like Dyna Mask \cite{Crabbe2021ExplainingTS}, and Extremal Mask \cite{enguehard2023learning} focus on \textbf{learning the masks} to better perturb the input features. Model-based saliency methods, such as \citep{Kaji2019AnAB, lim2021temporal, islam2023interpreting}, use the model architecture e.g. attention layers, to generate importance scores. ContraLSP \cite{liu2024explaining}, proposed a contrastive learning method to learn locally sparsed perturbation. TIMEX \cite{queen2024encoding} trained interpretable surrogates to learn stable explanations from latent space. However, TIMEX is limited to classification and assumes access to latent pretrained space. These methods overall have performed very well on synthetic data or real-world classification tasks. However, they \textbf{require training another model to interpret the target model}, which adds additional complexity. These are \textbf{not benchmarked for regression tasks} and often include algorithm design exclusively for classification \cite{queen2024encoding}. Also, \textbf{heavily uses simple RNN or LSTM baselines which are not state-of-the-art time series models}.


TSR \cite{ismail2020benchmarking} improved interpretation by considering the temporal dimension separately. However, this comes with a heavy computational cost and was not benchmarked on real-world time series data. Feature Importance in Time, FIT \cite{Tonekaboni2020WhatWW} conditioned on the last input time step to calculate the distribution shift using KL-divergence of the predicted probability, but only supports classification tasks. WinIT \cite{leung2022temporal} proposed a sliding window-based approach to calculate delayed feature importance for classification tasks.

So we need an interpretation method that is generalized for classification and regression, considers the time dependency of input data, and does not require training another model. In this work, we achieve this by, efficiently considering the time dependency, using a simple masking technique, and developing a novel framework that allows comparing these methods with SOTA time series models. 

\section{Windowed Temporal Saliency Rescaling \label{sec:WinTSR}}

\subsection{Problem Statement \label{sec:problem_statement}}

We consider a multivariate multi-horizon time series setting with length $T$, the number of input features $J$, target outputs $O$, and total $N$ instances. $X_{j,t} \in \textbf{R}^{J \times T}$ is the input feature $j$ at time $t \in \{0, \cdots, T-1\}$. Past information within a fixed look-back window $L$ is used to forecast for the next $\tau_{max}$ time steps or the target class. The target output at time $t$ is $y_t$. The black-box model $f$ is defined as,

\begin{equation}
\begin{aligned}
   \hat{y}_{t} & = f(X_t)  \\
 \text{where, } X_t &= x_{t-(L-1):t} = [x_{t-(L-1)}, x_{t-(L-2)}, \cdots, x_t]\\
&= \{ x_{j, l, t}\}, ~ j \in \{1, \cdots, J\}, ~ l \in \{1, \cdots, L\}
\end{aligned}
\end{equation}

where $\hat{y}_{t}$ is the predicted class or the forecast at $\tau \in \{1, \cdots, \tau_{max}\}$ time steps in the future. $ X_t$ is the input slice at time $t$ of length $L$. 
An input feature at position $(j, l)$ in the full input matrix at time step $t$ is denoted as $x_{j, l, t}$. 

The interpretation goal is to calculate the importance matrix, $\phi_t = \{ \phi_{j, l, t} \}$ for each output $o \in O$ or prediction horizon $\tau \in \{1, \cdots, \tau_{max}\}$. This is a matrix of size $O \times J \times L$ for classification and  $O \times \tau_{max} \times J \times L$ for regression. We find the relevance of the feature $x_{j, l, t}$ by masking it in the input matrix $X_t$ and measuring the model output change,

\begin{equation}
    \mathbf{\boldsymbol{\phi_{j, l, t}} = \mathbf{distance\_score} (f(X_t), ~ f(X_t ~\text{\textbackslash}~ x_{j, l, t}))}
\end{equation}

where $X_t ~\text{\textbackslash}~ x_{j, l, t}$ is the input after masking/perturbing feature $x_{j, l, t}$. The \textit{distance\_score} can vary based on interpretation methods and prediction tasks. For example, l1-norm for regression and l1-norm or KL-divergence for classification. 

\begin{algorithm}[!htb]
\caption{WinTSR: Windowed Temporal Saliency Rescaling}\label{alg:tsr}
\textbf{Given:} Input $X_t$, a time series model $f()$, look-back window $L$, number of features $J$. \\
\textbf{Output:} Feature importance matrix $\phi_t$\\~

baseline = $feature\_generator()$\\

\For{$l \gets 0$ to L}{
    \# Mask all features at time $l$ \\
    $X_t  ~\backslash~ x_{: , l , t} = baseline_{:, l, t}$ \\
    \# Compute Time-Relevance Score \\ 
    \hl{$\Delta^{time}_{l , t} = |f(X_t) -  f(X_t  ~\backslash~ x_{: , l , t} )| $}
}

\# to stabilize the temporal scaling
normalize $\Delta^{time}_{: , t}$\\

\For{$l \gets 0$ to L}{
\For{$j \gets 0$ to J}{
            \# Mask feature $j$ at lookback $l$\\ 
            $X_t  ~\backslash~ x_{j , l , t} = baseline_{j ,l , t}$\\
            \# Compute Feature-Relevance \\
            \hl{$\Delta^{feature}_{j, l, t} = |f(X_t) - f(X_t  ~\backslash~ x_{j, l, t} )| $} \\
        \# Compute final importance \\
        \hl{$\phi_{j, l, t} = \Delta^{feature}_{j, l, t} \times \Delta^{time}_{l, t}$}
    }
}
\Return $\phi$
\end{algorithm}

\subsection{Our Approach}

We propose \textit{Windowed Temporal Saliency Rescaling (WinTSR)} across the input time window to calculate the importance score matrix $\phi_t$ at a time $t$. Our method differs from previous approaches by accounting for the importance of a feature observation in a multi-horizon setting over multiple windows containing the target time step. The details are in the following Algorithm \ref{alg:tsr}. The method returns an importance matrix $\phi$ which is later used to evaluate the interpretation performance.

For a distance score, we calculate the simple `L1 distance` between the original and perturbed prediction for classification and regression. Unlike TSR, which uses the `L1 distance` between the original and perturbed importance matrix returned from another interpretation method. This significantly improves our run time compared to TSR and removes the dependency on a second interpretation method. WinIT perturbs all values in a sliding window with and without the target feature, then finds the feature importance by subtracting them. Since we perturb only the individual feature, this reduces the computation overhead of perturbing a range of features and removes the dependency of choosing a best-fit sliding window. 

The time relevance score enables us to skip less important time steps to speed up the computation similar to \cite{ismail2020benchmarking}. TSR uses the first feature value for masking, while WinIT uses feature augmentation (generated from past features). We generated random values from a normal distribution for masking and the input features are already normalized during the data pre-processing period.

\section{Experimental Setup \label{sec:exp_setup}}

We compare WinTSR to ten recent local and post-hoc interpretation methods. We evaluate them with five state-of-the-art deep learning time series models across three datasets for classification and regression tasks. 

\subsection{Datasets \label{sec:datasets}}

We use the datasets shown in Table \ref{tab:dataset}. Electricity and Traffic datasets contain the electricity consumption rate and traffic occupancy over the past hours respectively. The task is to forecast those values for the next 24 hours based on past observations and time-encoded features from the last 96 hours. The MIMIC-III dataset contains patient info and lab reports from a hospital. The goal is to predict whether the patient will die during their hospital stay, based on the patient demographic info and lab reports over the last 48 hours. This is a private dataset but easy to apply for access and popularly used in related works \cite{leung2022temporal, enguehard2023learning}. Details on the datasets and features are in Appendix \ref{app:dataset}. The input values are standard normalized. The datasets are split into train validation test sets using the 8:1:1 ratio. The best model by validation loss is used in testing and for the rest of the experiments. 

\begin{table}[!htb]
\small
    \centering
    \caption{Dataset descriptions with sample size. Each data has an hourly frequency.}

    \begin{adjustbox}{max width=\linewidth}
        \begin{tabular}{|l|c|c|c|c|l|} \hline
        \textbf{Dataset} & \textbf{Feature} & \textbf{Size} & \textbf{Window} & \textbf{Output} & \textbf{Task}\\ \hline
         Electricity &  5   & 26.2k & 96 & 24 & Regression \\
         Traffic &  5 & 20.7k & 96 & 24 & Regression \\
         MIMIC-III & 32   & 22.9k & 48 & 2 & Classification \\ \hline
    \end{tabular}
    \end{adjustbox}
    \label{tab:dataset}
\end{table}

\subsection{Models \label{sec:models}}

We use five neural network architecture groups (Linear, CNN, RNN, Transformer, and LLM) for our experiment. Multiple models are chosen to generalize the proposed method across different network architectures. We show how the same interpretation method impacts different models. A complete list of available models in our framework is given in Appendix \ref{sec:available_models}. We selected these five models based on their state-of-the-art performance in their respective architecture. These models are : (1) \textbf{DLinear} \citep{Zeng2022AreTE} - Linear, (2) \textbf{SegRNN} \citep{lin2023segrnn} - Recurrent Neural Network (RNN), (3) \textbf{MICN} \citep{wang2023micn} - Convolutional Neural Network (CNN), and (4) \textbf{iTransformer} \citep{liu2024itransformer} - Transformer, and (5) \textbf{CALF} \cite{liu2024taming} - A recent pretrained LLM model for generalized time series forecasting using cross-modal fine-tuning. 

\begin{table}[!htbp]
    \centering
    \caption{Test results. The best results are in bold.}
    \resizebox{.49\textwidth}{!}{
    \begin{tabular}{|p{1.4cm}|p{1cm}|wc{1cm}wc{.9cm}wc{1.1cm}wc{1cm}wc{1cm}|} 
        \hline
        \textbf{Dataset} &  \textbf{Metric} & \textbf{DLinear} & \textbf{MICN} & \textbf{SegRNN} & \textbf{iTrans.} & \textbf{CALF}\\ \hline
        \multirow{2}{*}{Electricity} & MAE & 0.36 & 0.39 & 0.27 & \textbf{0.24} &  0.28\\
         & MSE & 0.25 & 0.29 & 0.14 & \textbf{0.11} & 0.15 \\ \hline
        \multirow{2}{*}{Traffic} & MAE & 0.36 & 0.27 & 0.30 & \textbf{0.24} & 0.32 \\
         & MSE & 0.28 & 0.20 & 0.25 & \textbf{0.17} & 0.24 \\ \hline
        \multirow{2}{*}{MIMIC-III} & Acc & 0.90 & 0.90 & 0.90 & \textbf{0.91}& \textbf{0.91} \\
         & AUC & 0.77 & 0.77 & 0.78 & \textbf{0.82} & 0.77 \\ \hline
    \end{tabular}
    }
    \label{tab:test_results}
\end{table}

We follow the Time-Series-Library \cite{wu2023timesnet} \footnote{https://github.com/thuml/Time-Series-Library} implementation of these models. This ensures we follow the latest benchmarks. Table \ref{tab:test_results} shows the iTransformer model performs best across all cases. Details of the hyperparameters are listed in Appendix \ref{sec:hyperparameter}.

\subsection{Interpretation Methods\label{sec:interpretation_methods}}
We use the following post-hoc interpretation analysis methods for comparison in this work: (1) Feature Ablation (FA, \citet{Suresh2017ClinicalIP}) (2) Augmented Feature Occlusion (AFO, \citet{Tonekaboni2020WhatWW}) (3) Feature Permutation (FP, \citet{molnar2020interpretable}) (4) Integrated Gradients (IG, \cite{Sundararajan2017AxiomaticAF}) (5) Gradient Shap (GS, \citet{lundberg2017unified}) (6) Dyna Mask (DM, \citet{Crabbe2021ExplainingTS}) (7) Extremal Mask \citet{enguehard2023learning} (8) Windowed Feature Importance in Time (WinIT, \citet{leung2022temporal}) (9) Temporal Saliency Rescaling (TSR, \citet{ismail2020benchmarking}), and (10) Contrastive and Locally Sparse Perturbations (ContraLSP, \citet{liu2024explaining}). 

We choose them based on versatility across different model architectures and tasks. Captum \cite{kokhlikyan2020captum} \footnote{https://captum.ai/} and Time Interpret \cite{enguehard2023time} \footnote{https://josephenguehard.github.io/time\_interpret} libraries were used to implement the interpretation methods. Unlike \cite{enguehard2023time}, which runs the methods on the CPU, we implemented our framework to run all methods with GPU, thus increasing the interpretation speed. The baselines to mask inputs were randomly generated from the normal distribution, the raw inputs were also normalized. We excluded the methods which are classification only (e.g. FIT) or no public implementation is not available (e.g. CGS-Mask). For TSR, we used the best combination in their work (TSR with Integrated Gradients and $\alpha=0.55$).

\subsection{Evaluating Interpretation\label{sec:interpretation}}

\begin{figure}[!htb]
\centering
\includegraphics[width=0.47\textwidth]{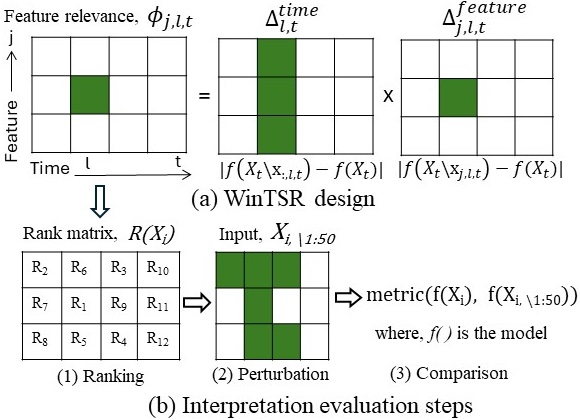}
\caption{An illustration of (a) our proposed WinTSR method and (b) evaluating interpretation without ground truth. The above example is about calculating comprehensiveness at k=50\% with three steps. }
\label{fig:indentify_feature}
\end{figure} 

We follow \cite{ozyegen2022evaluation, turbe2023evaluation} to evaluate interpretation when no interpretation ground truth is present. Figure \ref{fig:indentify_feature} (b) briefly illustrates the evaluation framework. The steps are:

\begin{enumerate}
    \item[1)] Sort relevance scores $\phi(X)$ so that $R_e(X_{i, t})$ is the $e^{th}$ element in the ordered rank set $\{R_e(x_{i, t})_{e=1}^{L \times N}$. Here $L$ is the look-back window and $N$ is the number of features. 
    \item[2)] Find top $k\% ~ (\text{we used, } k \in \{5, 7.5, 10, 15\})$ features in this set, where $\mathbf{R}(x_{i, t}) \in \{\mathbf{R}_e(x_{i, t})\}^k_{e=1}$. Mask these top features or every other feature in the input. 
    \item[3)] Calculate the change in the model's output to the original output using different metrics. We use the AUC drop for classification \cite{leung2022temporal} and Mean Absolute Error (MAE) for regression.
\end{enumerate}

\begin{table*}[!htb]
\small
    \centering
    \caption{\textbf{Interpretation benchmark results}, higher comprehensiveness, and lower sufficiency are better. For each dataset, model, and evaluation method, the best results are in bold and the second-best results are underlined. \textit{'-'} means results not run due to long runtime.}
    \small
    \begin{subtable}{\linewidth}
    \centering
    \captionsetup{skip=0pt}
    \caption{\textbf{Electricity} (AOPCR of MAE)}
\resizebox{.90\textwidth}{!}{    
    \begin{tabular}   {|p{1.4cm}|R{1cm}R{0.8cm}R{.8cm}R{1.1cm}R{.8cm}|R{1cm}R{0.8cm}R{.8cm}R{1.1cm}R{.8cm}|C{1.9cm}|} \hline
    
        \multirow{2}{=}{\textbf{Attribution Method}} & \multicolumn{5}{c|}{\textbf{Comprehensiveness} ($\pmb{\uparrow}$)} & \multicolumn{5}{c|}{\textbf{Sufficiency} ($\pmb{\downarrow}$)} & \textbf{Rank}\\ \cline{2-11}
        & \textbf{DLinear} & \textbf{MICN} & \textbf{SegRNN} & \textbf{iTrans.} & \textbf{CALF} & \textbf{DLinear} & \textbf{MICN} & \textbf{SegRNN} & \textbf{iTrans.} & \textbf{CALF} & \textbf{ (Avg  $\pm$  Std)}\\ \hline
FA & \underline{10.7} & 12.0 & 11.0 & 10.7 & 10.6 & 12.4 & \underline{11.1} & 13.4 & \underline{16.3} & \underline{16.4} & 2(2.6 $ \pm $ 0.5) \\
AFO & 10.0 & 11.1 & 10.2 & 9.1 & 9.3 & 13.0 & 12.1 & 14.0 & 17.3 & 17.0 & 3(4.1 $ \pm $ 0.7) \\
FP & 7.0 & 9.4 & 7.9 & 7.0 & 7.6 & 16.0 & 14.1 & 15.6 & 19.3 & 18.9 & 10(8.2 $ \pm $ 1.1) \\
IG & 10.7 & \textbf{13.7} & \underline{11.8} & \textbf{16.1} & \textbf{14.7} & 19.0 & 16.3 & 15.1 & 24.8 & 20.6 & 6(5.7 $ \pm $ 4.6) \\
GS & 7.7 & 10.3 & 9.1 & 8.2 & 9.6 & 17.4 & 14.6 & 15.3 & 20.4 & 19.7 & 8(7.4 $ \pm $ 2.0) \\
DM & 3.6 & 7.1 & 3.9 & 3.9 & 4.1 & 16.6 & 14.5 & 16.9 & 20.5 & 20.3 & 11(10.2 $ \pm $ 0.9) \\
EM & 8.7 & 9.7 & 4.8 & 5.4 & 8.3 & 13.6 & 13.2 & 16.6 & 19.2 & 18.0 & 8(7.4 $ \pm $ 1.8) \\
WinIT & 10.6 & 9.6 & 9.9 & 8.8 & 9.0 & 13.5 & 13.7 & 14.6 & 17.4 & 17.7 & 5(5.5 $ \pm $ 1.4) \\
TSR & 8.4 & 11.7 & 7.7 & 7.7 & - & \underline{11.9} & 14.1 & \textbf{12.0} & 16.4 & - & 4(5.0 $ \pm $ 2.8) \\
ContraLSP & 7.6 & 9.7 & 4.6 & 6.0 & 8.5 & 13.8 & 12.5 & 16.3 & 18.6 & 17.4 & 7(7.3 $ \pm $ 2.1) \\

\rowcolor{gray!50}  
WinTSR & \textbf{11.8} & \underline{12.7} & \textbf{12.7} & \underline{12.7} & \underline{13.0} & \textbf{11.3} & \textbf{9.8} & \underline{12.1} & \textbf{14.5} & \textbf{14.3} & \textbf{1(1.4 $ \pm $ 0.5)} \\
\hline
    \end{tabular}
}    
\end{subtable}

\vspace{1ex}

\begin{subtable}{\linewidth}
    \centering
    \captionsetup{skip=0pt}
    \caption{\textbf{Traffic} (AOPCR of MAE)}
\resizebox{.90\textwidth}{!}{        
    \begin{tabular}   {|p{1.4cm}|R{1cm}R{0.8cm}R{.8cm}R{1.1cm}R{.8cm}|R{1cm}R{0.8cm}R{.8cm}R{1.1cm}R{.8cm}|C{1.9cm}|} \hline
    
        \multirow{2}{=}{\textbf{Attribution Method}} & \multicolumn{5}{c|}{\textbf{Comprehensiveness} ($\pmb{\uparrow}$)} & \multicolumn{5}{c|}{\textbf{Sufficiency} ($\pmb{\downarrow}$)} & \textbf{Rank}\\ \cline{2-11}
        & \textbf{DLinear} & \textbf{MICN} & \textbf{SegRNN} & \textbf{iTrans.} & \textbf{CALF} & \textbf{DLinear} & \textbf{MICN} & \textbf{SegRNN} & \textbf{iTrans.} & \textbf{CALF} & \textbf{ (Avg  $\pm$  Std)}\\ \hline
FA & 10.4 & \underline{10.1} & 13.4 & 9.3 & 12.4 & 22.4 & 16.4 & \underline{17.9} & 22.8 & \underline{22.3} & 2(2.8 $ \pm $ 0.6) \\
AFO & 8.5 & 8.5 & 12.1 & 7.6 & 9.4 & 23.6 & 18.2 & 19.7 & 24.6 & 24.3 & 5(4.9 $ \pm $ 1.0) \\
FP & 7.5 & 7.1 & 10.4 & 7.2 & 10.0 & 24.8 & 19.4 & 20.7 & 24.6 & 24.8 & 7(6.4 $ \pm $ 1.1) \\
IG & \underline{10.9} & 9.4 & 10.8 & \textbf{13.4} & \textbf{13.7} & 24.1 & 18.1 & 21.2 & 24.8 & 23.7 & 4(3.8 $ \pm $ 2.1) \\
GS & 9.5 & 8.5 & 9.2 & 8.3 & 11.5 & 24.2 & 18.5 & 21.9 & 24.8 & 24.3 & 6(5.7 $ \pm $ 1.3) \\
DM & 3.5 & 4.4 & 4.1 & 3.8 & 4.0 & 26.3 & 21.4 & 23.9 & 27.9 & 28.1 & 11(10.6 $ \pm $ 0.5) \\
EM & 5.8 & 5.6 & 6.5 & 4.0 & 8.3 & 24.1 & 20.6 & 22.5 & 27.0 & 25.2 & 9(8.3 $ \pm $ 1.1) \\
WinIT & 7.3 & 6.1 & 11.7 & 6.3 & 7.2 & 25.3 & 21.1 & 22.3 & 25.7 & 26.5 & 8(8.0 $ \pm $ 1.2) \\
TSR & 8.7 & 7.0 & \underline{13.5} & 9.6 & - & \textbf{15.6} & \textbf{11.9} & 19.6 & \underline{22.6} & - & 3(3.0 $ \pm $ 2.1) \\
ContraLSP & 4.2 & 2.9 & 4.6 & 4.0 & 6.1 & 26.1 & 22.5 & 23.0 & 27.1 & 23.9 & 10(9.5 $ \pm $ 2.0) \\
\rowcolor{gray!50}  
WinTSR & \textbf{11.2} & \textbf{10.8} & \textbf{14.3} & \underline{10.8} & \underline{13.1} & \underline{21.2} & \underline{15.5} & \textbf{16.5} & \textbf{21.3} & \textbf{20.8} & \textbf{1(1.4 $ \pm $ 0.5)} \\
\hline
    \end{tabular}
}
\end{subtable}

\vspace{1ex}
\begin{subtable}{\linewidth}
    \centering
    \captionsetup{skip=0pt}
    \caption{\textbf{MIMIC-III} (AOPC of AUC drop)}
\resizebox{.90\textwidth}{!}{        
    \begin{tabular}{|p{1.4cm}|R{1cm}R{0.8cm}R{.8cm}R{1.1cm}R{.8cm}|R{1cm}R{0.8cm}R{.8cm}R{1.1cm}R{.8cm}|C{1.9cm}|} \hline
    
        \multirow{2}{=}{\textbf{Attribution Method}} & \multicolumn{5}{c|}{\textbf{Comprehensiveness} ($\pmb{\uparrow}$)} & \multicolumn{5}{c|}{\textbf{Sufficiency} ($\pmb{\downarrow}$)} & \textbf{Rank}\\ \cline{2-11}
        & \textbf{DLinear} & \textbf{MICN} & \textbf{SegRNN} & \textbf{iTrans.} & \textbf{CALF} & \textbf{DLinear} & \textbf{MICN} & \textbf{SegRNN} & \textbf{iTrans.} & \textbf{CALF} & \textbf{(Avg $\pm$  Std)}\\ \hline

FA & \underline{0.61} & \underline{0.50} & 0.51 & 0.46 & 0.82 & 0.49 & 0.46 & \underline{0.30} & 0.49 & 0.83 & 2(3.6 $ \pm $ 1.6) \\
AFO & 0.53 & 0.46 & 0.45 & 0.42 & 0.82 & \underline{0.45} & \underline{0.43} & \textbf{0.27} & 0.48 & 0.83 & 3(4.8 $ \pm $ 2.6) \\
FP & 0.55 & 0.46 & 0.31 & 0.42 & 0.83 & 0.64 & 0.54 & 0.40 & 0.57 & 0.85 & 7(6.6 $ \pm $ 2.0) \\
IG & 0.51 & 0.48 & 0.28 & 0.42 & 0.83 & 0.61 & 0.54 & 0.48 & 0.51 & 0.84 & 8(6.9 $ \pm $ 1.9) \\
GS & 0.55 & 0.49 & 0.24 & 0.41 & \underline{0.84} & 0.64 & 0.57 & 0.45 & 0.55 & 0.87 & 9(7.5 $ \pm $ 2.9) \\
DM & 0.48 & 0.43 & 0.28 & 0.40 & 0.72 & 0.56 & 0.56 & 0.54 & 0.54 & \underline{0.78} & 11(8.2 $ \pm $ 2.6) \\
EM & 0.53 & 0.40 & 0.25 & 0.43 & 0.73 & 0.55 & 0.55 & 0.56 & 0.58 & \textbf{0.74} & 10(7.8 $ \pm $ 3.0) \\
WinIT & 0.50 & 0.45 & 0.46 & 0.40 & 0.83 & 0.53 & 0.44 & 0.31 & 0.51 & 0.84 & 6(6.3 $ \pm $ 2.8) \\
TSR & \textbf{0.80} & \textbf{0.90} & \textbf{0.79} & \textbf{0.83} & - & 0.90 & 0.95 & 0.87 & 0.91 & - & 5(6.0 $ \pm $ 5.3) \\
ContraLSP & 0.59 & 0.44 & 0.26 & 0.44 & 0.83 & 0.48 & 0.45 & 0.48 & \underline{0.47} & 0.82 & 4(4.9 $ \pm $ 2.6) \\
\rowcolor{gray!50}  
WinTSR & 0.56 & 0.50 & \underline{0.52} & \underline{0.48} & \textbf{0.85} & \textbf{0.44} & \textbf{0.42} & 0.32 & \textbf{0.47} & 0.83 & \textbf{1(2.4 $ \pm $ 1.5)} \\
\hline

\end{tabular}
}
\end{subtable}

\vspace{1ex}
{\raggedright \textbf{Abbreviations}: \textit{AOPC}: Area over the perturbation curve for classification, \textit{AOPCR}: Area over the perturbation curve for regression, \textit{FA}: Feature Ablation,  \textit{AFO}: Augmented Feature Occlusion, \textit{FP}: Feature Permutation, \textit{IG}: Integrated Gradients, \textit{GS}: Gradient Shap, \textit{DM}: Dyna Mask, \textit{EM}: Extremal Mask, \textit{WinIT}: Windowed Feature Importance in Time, \textit{TSR}: Temporal Saliency Scaling with Integrated Gradients,\textit{ContraLSP}: Contrastive and Locally
Sparse Perturbation, \textit{WinTSR}: Windowed Temporal Saliency Rescaling. \par}

\label{table:int_benchmark}
\end{table*}

\begin{figure*}[!htb]
\centering
\subfloat[Electricity]{
\includegraphics[width=0.90\linewidth]{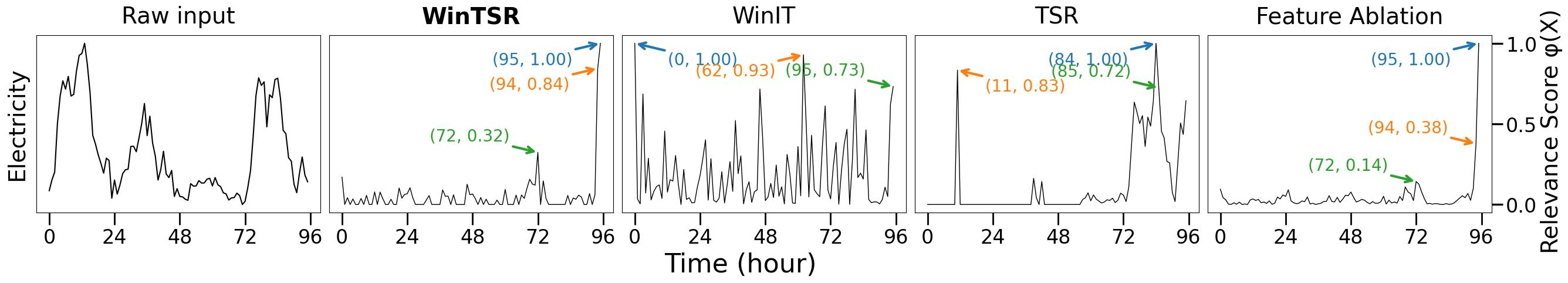}}
\hspace{.1ex}
\subfloat[Traffic]{
\includegraphics[width=0.90\linewidth]{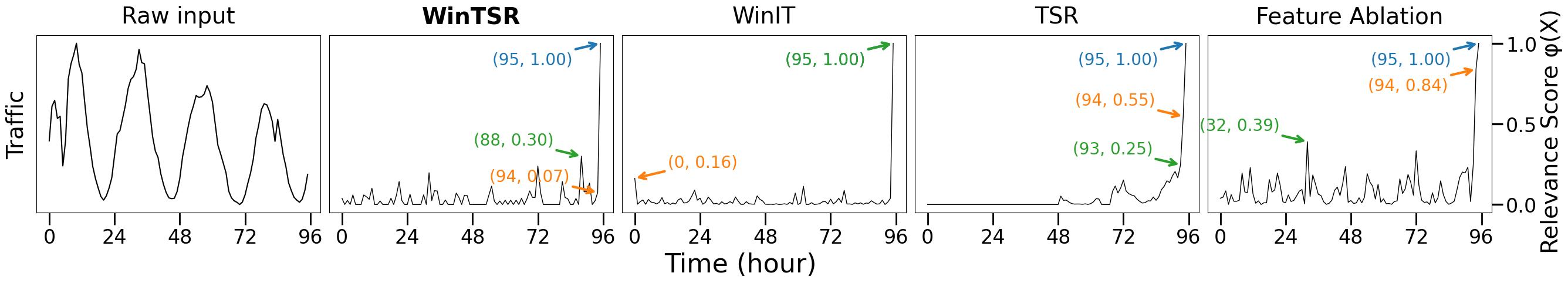}}
\hspace{0.1ex}

\subfloat[MIMIC-III]{
\includegraphics[width=0.96\linewidth]{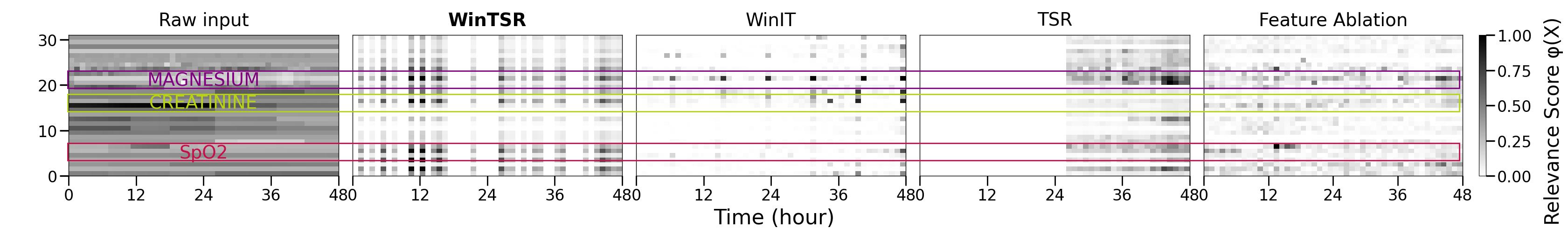}
}

\caption{Interpreted relevance scores (when $\tau=1$ or $o=1$) for an arbitrarily chosen example from each dataset, some selected methods, and the iTransformer model. For MIMIC-III the overall most important features of each method are annotated. Both electricity and traffic features show high importance for more recent features and a 24-hour pattern for most methods.}
\label{fig:visualization}
\end{figure*} 

\citet{deyoung2019eraser} proposed to measure the \textit{comprehensiveness} and \textit{sufficiency} to ensure the faithfulness of the explained rationales. Which are similar to the precision and recall from \citet{ismail2020benchmarking}. (1) \textit{Comprehensiveness}: Were all features needed to make a prediction selected? Once important features are masked, the model should be less confident in its prediction. (2) \textit{Sufficiency:} Are the top feature sufficient to make the prediction? This is achieved by masking all features except the top $k\%$. In summary, \textbf{the higher the comprehensiveness loss and the lower the sufficiency loss the better}. We define the set of top $k\%$ relevant features selected by the interpretation method for the $i$-th input $X_i$ as $X_{i, 1:k}$, the input after removing those features as $X_{i, ~\text{\textbackslash} 1:k}$. Then these two terms are calculated as: 
 \begin{equation}
 \begin{aligned}
     \text{Comp.} &= \text{evaluation\_metric}( f(X_i), ~f(X_{i, ~\text{\textbackslash} 1:k}  )) \\
      \text{Suff.}  & = \text{evaluation\_metric}( f(X_i), ~f(X_{i, 1:k}  ))
 \end{aligned}
\end{equation}

For $K$ bins of top $k\%$ features (we use top 5\%, 7.5\%, 10\%, and 15\% features, hence $K=4$.), the aggregated comprehensiveness score \citep{deyoung2019eraser} for the classification task is called the "Area Over the Perturbation Curve" (\textit{AOPC}, Equation \ref{eqn:aopc}). For AUC drop, this will calculate the drop for each output class $o$ after masking top $k\%$ features for each $k \in K$, then calculate the average drop. 

Similarly for regression, \cite{ozyegen2022evaluation} defined the "Area Over the Perturbation Curve for Regression" (\textit{AOPCR}, Equation \ref{eqn:aopcr}). For MAE, it calculates the change in prediction for each output $o$ and prediction horizon $\tau$ by masking top $k\%$ features for each $k \in K$ then takes the average. AOPC and AOPCR for sufficiency are calculated similarly after replacing $X_{i,  ~\text{\textbackslash} 1:k}$ with $X_{i, 1:k}$. 

\begin{equation}
 \label{eqn:aopc}
AOPC = \frac{\displaystyle\sum_{o,k}^{O, K} \text{metric}(f(X_i)_o, f(X_{i,  ~\text{\textbackslash} 1:k})_o)}{K \times O} 
\end{equation}

\begin{equation}
 \label{eqn:aopcr}
    \begin{aligned}
    AOPCR = & \frac{\displaystyle\sum_{o, \tau, k}^{O, \tau_{max}, K}
     \text{metric}(f(X_i)_{o, \tau}, f(X_{i,  ~\text{\textbackslash} 1:k})_{o, \tau})}{K \times O \times \tau_{max}} 
\end{aligned}
\end{equation}

\section{Results \label{sec:results}}

This section shows the interpretation results and visualizations. Then discuss our time complexity and the effect of changing the lookback window.

\subsection{Benchmark Evaluation}

Table \ref{table:int_benchmark} shows the overall results. Our method performs the best or second best in most cases. This is consistent across different datasets and models. We ranked the methods for each dataset and model in terms of overall comprehensiveness and Sufficiency. Then we averaged the ranks in the rightmost columns and used them for the final rank. \textbf{WinTSR achieves the best average rank in each dataset}, 1(1.4 $\pm$ 0.5), 1(1.4 $\pm$ 0.05), and 1(2.4 $\pm$ 1.5) in the Electricity, Traffic, and MIMIC-III respectively. 

Integrated Gradient achieves the best results in a few cases for comprehensiveness in regression but fails in others. TSR performs significantly better for comprehensiveness in the MIMIC-III dataset, but its high sufficiency in the same dataset shows that the top features it selects are insufficient.  Feature Ablation method also consistently performed well and achieved 2nd rank overall. We also see the mask learner methods, in practice do not interpret the SOTA models well. 

\subsection{Visualizing Interpretation \label{sec:visualization}}

Visualizing the interpretations helps to understand their meaning. However, unlike images and texts, time series interpretations are harder to visualize and to verify intuitively. Here we 1) visualize persistent temporal patterns (trends present across the dataset) and 2) highlight the top features across time. Figure \ref{fig:visualization} shows the raw input feature (left) and the interpretation of these features (using normalized relevance/importance score). The MIMIC-III dataset is visualized with a heatmap due to many features (31), the other two datasets are shown with line plots. The interpretation results of the four selected methods are presented for comparison using the best-performing iTransformer model of 1st iteration.

The relevance scores shown here are for forecasting the target for the next hour ($\tau=1$) or predicting mortality class ($o=1$) for simplicity. Electricity and traffic features show a daily pattern, with the highest importance at the most recent time step ($t=96$) and the same time the previous day ($t=72$). Sometimes at the last peak. This means, that to predict the electricity consumption or traffic occupancy, past observations from recent times or the same daytime or last peak hour are important. For MIMIC-III the goal is to interpret which factors at which time were important in predicting the patient's death. Figure \ref{fig:visualization} (c) shows the top three points interpreted by the methods, where WinIT and TSR display the features important in the last 12 hours, whereas WinTSR and FA identify these features much earlier, within the first 12 hours, and then again around the last 12 hours. Temporal change of the important features is visible in WinTSR, WinIT, and TSR as they all consider temporal dependency.

\subsection{Time Complexity}
We summarize the run time information in Table \ref{tab:time_efficiency} for some selected methods, Appendix \ref{tab:time_efficiency_full} includes the complete results. WinTSR's time complexity is $O(L+ L \times J)$, where $L$ is the lookback window and $J$ is the number of features. The perturbation-based methods (FA, AFO, FP) have similar run-time efficiency $O(L \times J)$. WinIT has time complexity $O(L \times J)$, but since it needs to perturb a sliding window of feature each time, it is slower in practice. Gradient-based methods (IG, GS, DM) run the fastest. The \textit{TSR} method is the slowest since it repeatedly applies the \textit{IG} method across each time and feature column, then along the time axis to calculate the time relevance score. The time complexity is $O((L + L \times J) \times O_{IG})$ where $O_{IG}$ is the time complexity of the Integrated Gradient (IG) method. In practice, \textbf{WinTSR is around 32 to 367 times faster than TSR}.

\begin{table}[!htb]
\small
\centering
\caption{\textbf{Runtime (minutes)}: A comparison of the interpretation methods averaged across iterations (See the full comparison in Appendix \ref{tab:time_efficiency_full}). \textit{WinTSR} has a runtime similar to other perturbation-based methods, faster than \textit{WinIT}, and 32 to 367 times faster than \textit{TSR}. \textit{'-'} means results not run due to long runtime.}
\resizebox{.48\textwidth}{!}{
    \begin{tabular}{|L{2.2cm}|C{.4cm}C{.7cm}C{.5cm}C{1.3cm}C{1cm}|} \hline
      \textbf{Model} & \textbf{FA} & \textbf{WinIT} & \textbf{TSR} & \textbf{ContraLSP} & \textbf{WinTSR} \\ \hline
 DLinear (Linear)& 1.8 &  2.8 &  176 &  2.4 & \cellcolor{gray!50} 1.9 \\ 
  MICN (RNN)& 6.4 &  7.4 &  682 &  7.5 & \cellcolor{gray!50} 5.7 \\ 
 SegRNN (CNN)& 2.6 &  3.6 &  725 &  3.2 & \cellcolor{gray!50} 2.5 \\ 
 iTransformer (Trans.)& 5.2 &  6.2 &  1002 &  5.1 & \cellcolor{gray!50} 6.3 \\ 
 CALF (LLM)& 39.8 &  36.6 &  - &  48.7 & \cellcolor{gray!50} 43.3 \\ 
\hline
\end{tabular}
}
\label{tab:time_efficiency}

\end{table}

\subsection{Varying Lookback Window}

Since the lookback window size is an integral part of capturing temporal dependency, it is important to analyze the effect of changing the window size. By design, the WinIT method supports variable window length, where TSR and WinTSR compute over the whole training window size. We retrained the best-performing iTransformer model for different lookback windows and interpreted it by comparing the 3 window-based methods (WinIT, TSR, WinTSR), specifically on temporal dependency. The results are shown in Table \ref{tab:varying_lookback}. We reduced the lookback to 24-hour and 48-hour for Electricity and Traffic (original data have 96-hour lookback). For MIMIC-III, we varied the lookback to 24-hour and 36-hour since the original data had a 48-hour lookback. WinTSR performs best or 2nd best in most cases, showing its robustness across different input window sizes.

\begin{table}[!htb]
\small
    \centering
        \caption{Interpretation performance with different lookback windows for methods considering temporal dependency.}
    \resizebox{.48\textwidth}{!}{
    \begin{tabular}{|p{1.5cm}|l|wc{1.2cm}wc{1.1cm}|wc{1.2cm}wc{1.1cm}|}  \hline
    
    \multirow{2}{*}{\textbf{Dataset}} & \multirow{2}{*}{\textbf{Method}} & \multicolumn{2}{c|}{\textbf{Lookback = 24hr}} & \multicolumn{2}{c|}{\textbf{Lookback = 48hr}}\\ \cline{3-6}
    & & \textbf{Comp.} ($\pmb{\uparrow}$) & \textbf{Suff.} ($\pmb{\downarrow}$)& \textbf{Comp.} ($\pmb{\uparrow}$)& \textbf{Suff.} ($\pmb{\downarrow}$)\\ \hline
    \multirow{3}{*}{ Electricity} & WinIT & \underline{14.4} & 23.0 & \underline{11.5} & 17.9\\
 & TSR & 7.81 & \underline{21.5} & 8.15 & \underline{16.0} \\
 & \cellcolor{gray!35}WinTSR & \cellcolor{gray!35}\textbf{17.9} & \cellcolor{gray!35}\textbf{20.7} & \cellcolor{gray!35}\textbf{14.9} & \cellcolor{gray!35} \textbf{15.3}\\
\hline
\multirow{3}{*}{ Traffic} & WinIT & \underline{9.10} & \underline{25.5} & \underline{ 7.50} & 26.1\\
 & TSR & 6.10 & 25.9 & 7.49 & \underline{ 23.2}\\
 & \cellcolor{gray!35}WinTSR & \cellcolor{gray!35}\textbf{11.4 }& \cellcolor{gray!35}\textbf{23.3} & \cellcolor{gray!35}\textbf{11.3} & \cellcolor{gray!35}\textbf{22.7}\\
\hline

& & & & \multicolumn{2}{c|}{\textbf{Lookback = 36hr}} \\ \hline
\multirow{3}{*}{ MIMIC-III} & WinIT & 0.55 & \textbf{0.63} & 0.45 & \underline{0.55}\\
 & TSR & \textbf{0.87} & \underline{0.93} & \textbf{0.83} & 0.91 \\

 & \cellcolor{gray!35}WinTSR & \cellcolor{gray!35}\underline{0.59} & \cellcolor{gray!35} \textbf{0.63} & \cellcolor{gray!35} \underline{0.50} & \cellcolor{gray!35} \textbf{0.49}\\
\hline      
    \end{tabular}
    }
    \label{tab:varying_lookback}
\end{table}

\section{Conclusion and Future Work\label{sec:conclusion}}

In this paper, we present a novel local interpretation method "Windowed Temporal Saliency Rescaling" that explicitly accounts for the dynamic temporal nature of input data and explains their features' importance. Through extensive experiments and metric comparisons, our analysis 1) \textbf{shows WinTSR provides a more accurate interpretation of temporal dependencies among features}; 2) benchmarks different neural network models: DLinear (Linear), SegRNN (RNN), MICN (CNN), iTransformer (Transformer), and CALF (LLM). 3) compares with ten widely used interpretation methods; 4) presents an easy-to-use framework by combining a popular time series library with interpretation libraries. This framework enables the quantitative measurement of time series interpretation across many recent models and methods. For future work, we will identify higher-level patterns and trends in time series models by explicitly incorporating both spatial and temporal domains.

\section*{Acknowledgement}
This work is supported by NSF grant CCF-1918626  Expeditions: Collaborative  Research: Global Pervasive Computational Epidemiology.

\bibliography{aaai24}

\appendix
\section{Dataset and Features \label{app:dataset}}
\subsection{Electricity}
The UCI Electricity dataset \cite{misc_electricityloaddiagrams20112014_321} contains the consumption of 321 customers from 2012 to 2014. We aggregated it on an hourly level. This data has been used as a benchmark in many time series forecasting models \citep{wu2023timesnet, Zeng2022AreTE, ozyegen2022evaluation}. Following \cite{wu2021autoformer} we use the past 96 hours to forecast over the next 24 hours. And we added four time-encoded features: month, day, hour, and day of the week.

\subsection{Traffic}
The UCI PEM-SF Traffic Dataset 
describes the occupancy rate (with $y_t \in [0, 1]$) of 440 SF Bay Area freeways from 2015 to 2016. It is also aggregated on an hourly level. Following \cite{wu2021autoformer} we used a look-back window of 96 hours, a forecast horizon of 24 hours, and the 821st user as the target variable. We added four time-encoded features: month, day, hour, and day of the week.

\subsection{MIMIC-III Mortality}

\begin{table}[htb]
    \centering
    \small
    \caption{List of MIMIC-III dataset features.}
    \begin{tabular}{p{1.2cm}|p{5.8cm}} \hline
         \textbf{Features} & \textbf{Name} \\ \hline
         Static & Age, Gender, Ethnicity, ICU admission time\\ \hline
        Laboratory & ANION GAP, ALBUMIN, BICARBONATE, BILIRUBIN, 
            CREATININE, CHLORIDE, GLUCOSE, HEMATOCRIT, 
            HEMOGLOBIN, LACTATE, MAGNESIUM, PHOSPHATE, 
            PLATELET, POTASSIUM, PTT, INR, 
            PT, SODIUM, BUN, WBC \\\hline
        Vitality & HeartRate, DiasBP, SysBP, MeanBP,
        RespRate, SpO2, Glucose, Temp \\ \hline
    \end{tabular}
    \label{tab:mimic_iii_features}
\end{table}

A  multivariate real-world clinical time series dataset with a range of vital and lab measurements taken over time for over 40,000 patients \citep{Johnson2016MIMICIIIAF}. It is widely used in healthcare and medical AI-related research, and also in time series interpretation \cite{Tonekaboni2020WhatWW, leung2022temporal}. We follow the pre-processing procedure by \cite{leung2022temporal} to drop patients with missing information and then aggregate. Among the 22988 patient data left in the dataset, 2290 died during their hospital stay. We use the measurements hourly over 48 hours to predict patient mortality (whether the patient died). Table \ref{tab:mimic_iii_features} lists the clinical features used from the MIMIC-III patient dataset used in our experiments. There are four static features, twenty lab measurements, and eight vitality indicators. 

\section{Parameters and Notations \label{sec:hyperparameter}}

The model and training parameters are chosen following \cite{wu2023timesnet} for a consistent comparison with the state-of-the-art. Table \ref{tab:training_parameters} and \ref{tab:hyperparameters}  list the training parameters and the model hyperparameters used during our experiments. Table \ref{tab:notations} summarizes the notations mainly defined during the problem statement and interpretation evaluation framework. 

\begin{table}[htbp]
    \centering
    \caption{Training parameters.}
    \begin{tabular}{|p{1.6cm}|p{2.2cm}|p{1.6cm}|p{.8cm}|} \hline
         \textbf{Parameter} & \textbf{Value} & \textbf{Parameter} & \textbf{Value} \\ \hline
         epoch & 10 & optimizer & Adam \\
         dropout & 0.1 & batch size & 32 \\ 
         device & GPU & iterations & 3 \\
         loss & MSE or Cross Entropy &learning rate &  1e-3\\
        seed & 2024  & &  \\
         \hline
    \end{tabular}
    \label{tab:training_parameters}
\end{table}

\begin{table}[!htb]
\centering
\caption{Experiment configuration of the models. 
}
\resizebox{.47\textwidth}{!}{
\begin{tabular}{|p{1.7cm}|p{1.3cm}c|p{1.3cm}c|} \hline
     \multirow{2}{*}{\textbf{Model}} & \multicolumn{4}{c|}{\textbf{Hyper-parameters}} \\ \cline{2-5}
     & \textbf{Parameter} & \textbf{Value} & \textbf{Parameter} & \textbf{Value} \\ \hline
     DLinear & moving avg & 25 & &  \\\hline
     \multirow{2}{*}{MICN}  & encoders & 2 & kernel & (18, 12)  \\
     & decoders & 1 & dimension & 128 \\ \hline
     \multirow{2}{*}{SegRNN} & layers & 1 & dimension & 128 \\
     & seg\_len & 24 & & \\\hline
     \multirow{3}{*}{iTransformer}  & encoders & 2 & dimension & 128 \\ 
     & n\_heads & 4 & FCN dim & 256 \\
     & factor & 3 & & \\ \hline
     \multirow{3}{*}{CALF}  & gpt\_layers & 6 & dimension & 768 \\ 
     & tmax & 20 & lora\_alpha & 32 \\
     & lora\_dropout & 0.1 & & \\
     \hline
\end{tabular}
}    
\label{tab:hyperparameters}
\end{table}

\begin{table}[!htb]
    \centering
    \caption{Notations used in our work.}
    \begin{tabular}{p{1.3cm}|l} \hline
         \textbf{Notation} & \textbf{Meaning}  \\ \hline
         $T$ & time series length \\
         $f$ & model \\
         $X$ & input features \\
         $J$ & number of features \\
         $L$ & input lookback window \\
         $y$ & model output \\
         $ X_t \ $ & perturbed input \\
         $\hat{y}$ & model output on perturbed input \\
         $\phi$ & feature relevance matrix \\
         $R$ & rank matrix created from $\phi$ \\
         $O$ & number of output \\
         $\tau$ & output horizon \\
         $k$ & top \% of features \\ \hline
    \end{tabular}
    
    \label{tab:notations}
\end{table}

\begin{table*}[htb]
\centering
\small
\caption{\textbf{Runtime (minutes)}: The complete comparison of the interpretation methods averaged across iterations. \textit{WinTSR} has a runtime similar to other perturbation-based methods. While \textit{WinIT} is slightly slower and TSR is around 32-367 times slower.} 
   
    \begin{tabular}{|l|l|ccccccccccc|c|} \hline
    \textbf{Dataset} & \textbf{Model} & \textbf{FA} & \textbf{AFO} & \textbf{FP} & \textbf{IG} & \textbf{GS} & \textbf{DM} & \textbf{EM} & \textbf{WinIT} & \textbf{TSR} & \textbf{ContraLSP} & \textbf{WinTSR} \\ \hline
\multirow{4}{*}{Electricity} & DLinear& 1.83 & 1.97 & 1.91 & 2.29 & 1.48 & 1.44 & 1.36 & \cellcolor{gray!25} 2.8 & \cellcolor{gray!25} 176.3 & \cellcolor{gray!25} 2.4 & \cellcolor{gray!50} 1.9 \\ 
 & MICN& 6.44 & 6.56 & 6.52 & 7.09 & 4.59 & 4.46 & 4.45 & \cellcolor{gray!25} 7.4 & \cellcolor{gray!25} 682.2 & \cellcolor{gray!25} 7.5 & \cellcolor{gray!50} 5.7 \\ 
 & SegRNN& 2.64 & 2.78 & 2.71 & 2.84 & 2.01 & 1.96 & 1.89 & \cellcolor{gray!25} 3.6 & \cellcolor{gray!25} 725.4 & \cellcolor{gray!25} 3.2 & \cellcolor{gray!50} 2.5 \\ 
 & iTransformer& 5.22 & 5.36 & 5.32 & 4.57 & 3.72 & 3.62 & 4.68 & \cellcolor{gray!25} 6.2 & \cellcolor{gray!25} 1004 & \cellcolor{gray!25} 5.1 & \cellcolor{gray!50} 6.3 \\ 
 & CALF& 39.8 & 40.9 & 40.7 & 17.7 & 10.8 & 18.3 & 22.1 & \cellcolor{gray!25} 36.6 & \cellcolor{gray!25} - & \cellcolor{gray!25} 48.7 & \cellcolor{gray!50} 43.3 \\ 
\hline

\multirow{4}{*}{Traffic} & DLinear& 1.22 & 1.31 & 1.27 & 1.57 & 0.978 & 0.939 & 0.878 & \cellcolor{gray!25} 1.8 & \cellcolor{gray!25} 192.6 & \cellcolor{gray!25} 1.6 & \cellcolor{gray!50} 1.2 \\ 
 & MICN& 4.38 & 4.46 & 4.43 & 4.59 & 3.11 & 3.01 & 2.96 & \cellcolor{gray!25} 5.0 & \cellcolor{gray!25} 517.6 & \cellcolor{gray!25} 5.0 & \cellcolor{gray!50} 3.8 \\ 
 & SegRNN& 1.76 & 1.85 & 1.82 & 1.89 & 1.33 & 1.3 & 1.24 & \cellcolor{gray!25} 2.3 & \cellcolor{gray!25} 625.2 & \cellcolor{gray!25} 2.1 & \cellcolor{gray!50} 1.7 \\ 
 & iTransformer& 3.54 & 3.64 & 3.6 & 3.16 & 2.52 & 2.48 & 3.12 & \cellcolor{gray!25} 4.2 & \cellcolor{gray!25} 609.1 & \cellcolor{gray!25} 3.5 & \cellcolor{gray!50} 4.3 \\ 
 & CALF& 13.7 & 13.7 & 13.7 & 12.4 & 7.3 & 12.2 & 14.7 & \cellcolor{gray!25} 15.0 & \cellcolor{gray!25} - & \cellcolor{gray!25} 16.0 & \cellcolor{gray!50} 15.8 \\ 
\hline

\multirow{4}{*}{MIMIC-III} & DLinear& 1.85 & 2.23 & 2.05 & 0.361 & 0.3 & 0.461 & 0.439 & \cellcolor{gray!25} 6.8 & \cellcolor{gray!25} 176.0 & \cellcolor{gray!25} 4.0 & \cellcolor{gray!50} 4.4 \\ 
 & MICN& 6.92 & 7.31 & 7.12 & 0.85 & 0.617 & 0.839 & 0.911 & \cellcolor{gray!25} 12.0 & \cellcolor{gray!25} 476.2 & \cellcolor{gray!25} 5.7 & \cellcolor{gray!50} 7.4 \\ 
 & SegRNN& 3.82 & 4.71 & 4.69 & 0.772 & 0.483 & 0.594 & 1.03 & \cellcolor{gray!25} 7.9 & \cellcolor{gray!25} 373.3 & \cellcolor{gray!25} 4.6 & \cellcolor{gray!50} 5.9 \\ 
 & iTransformer& 5.21 & 5.61 & 5.41 & 0.617 & 0.511 & 0.706 & 0.9 & \cellcolor{gray!25} 10.2 & \cellcolor{gray!25} 249.3 & \cellcolor{gray!25} 3.5 & \cellcolor{gray!50} 6.4 \\ 
 & CALF& 102 & 103 & 102 & 7.77 & 3.99 & 26 & 38.9 & \cellcolor{gray!25} 106.2 & \cellcolor{gray!25} - & \cellcolor{gray!25} 19.8 & \cellcolor{gray!50} 60.2 \\ 
\hline
    \end{tabular}
     \label{tab:time_efficiency_full}
\end{table*}

\section{Interpretation methods} \label{app:interpretation}
The following describes the interpretation methods we have compared within this paper. 

\begin{enumerate}
    \item \textbf{\textit{Feature Ablation (FA):}}  The difference in output after replacing each feature with a baseline. \cite{Suresh2017ClinicalIP}. 
    \item \textbf{\textit{Augmented Feature Occlusion (AFO):}} \citet{Tonekaboni2020WhatWW} ablated the input features by sampling counterfactuals from the bootstrapped distribution. 
    \item \textbf{\textit{Feature Permutation (FP):}}  Permutes the input feature values within a batch and computes the difference between original and shuffled outputs \cite{molnar2020interpretable}. 
    \item \textbf{\textit{Integrated Gradients (IG):}} \citet{Sundararajan2017AxiomaticAF} assigned an importance score to each input feature by approximating the integral of gradients of the model’s output to the inputs.
    \item \textbf{\textit{Gradient Shap (GS):}} \citet{lundberg2017unified} approximated SHAP values by computing the expectations of gradients by randomly sampling from the distribution of baselines/references. 
    \item \textbf{\textit{Dyna Mask (DM):}} \citet{Crabbe2021ExplainingTS} learned masks representing feature importance. 
    \item \textbf{\textit{Extremal Mask (EM):}} \citet{enguehard2023learning} improved the static perturbation from Dyna Mask by learning not only masks but also associated perturbations.
    \item \textbf{\textit{Windowed Feature Importance in Time (WinIT):}}  \citet{leung2022temporal} explicitly accounted for the temporal dependence among observations of the same feature by summarizing its importance over a lookback window.
    \item \textbf{\textit{Temporal Saliency Rescaling (TSR):}} \citet{ismail2020benchmarking} proposed to separate the temporal dimension when calculating feature importance and rescaling it.
    \item \textbf{\textit{ContraLSP:}} \citet{liu2024explaining} designed a contrastive learning-based masking method to learn locally sparse perturbations for better explaining feature relevance with and without top important features.
\end{enumerate}


\section{Time Complexity}
Table \ref{tab:time_efficiency_full} shows the full run time comparison between the interpretation methods.

\section{Available Models \label{sec:available_models}}

Our framework currently includes the following time series foundation models: 
\begin{enumerate}
    \item \textit{CALF}: Aligns LLMs for time series forecasting with cross-modal fine-tuning \cite{liu2024taming}.
    \item \textit{TimeLLM}: Reprograms LLMs and its tokenization for better forecasting \cite{jin2023time}.
    \item \textit{GPT4TS}: Generalizes pretrained LLMs (GPT-2, Bert) for time series \cite{zhou2023one}.
\end{enumerate}

We include the following transformer-based and other recent time series models in our proposed framework:   
\begin{enumerate*}[label=(\arabic*)]
    \item \textit{Transformer}, \citet{vaswani2017attention}
    \item \textit{Reformer}, \citet{kitaev2020reformer}
    \item \textit{Informer}, \citet{zhou2021informer}
    \item \textit{Autoformer}, \citet{wu2021autoformer}
    \item \textit{FiLM}, \citet{zhou2022film}
    \item \textit{Pyraformer}, \citet{liu2022pyraformer}
    \item \textit{FEDformer}, \citet{zhou2022fedformer}
    \item \textit{Non-stationary Transformer}, \citet{liu2022non}
    \item \textit{ETSformer}, \citet{woo2022etsformer}
    \item \textit{LightTS}, \citet{zhang2022less} 
    \item \textit{PatchTST}, \citet{nie2022time}
    \item \textit{DLinear}, \citet{Zeng2022AreTE}
    \item \textit{TimesNet}, \citet{wu2022timesnet}
    \item \textit{TiDE}, \citet{das2023long}
    \item \textit{MICN}, \citet{wang2023micn} 
    \item \textit{SegRNN}, \citet{lin2023segrnn}
    \item \textit{Crossformer}, \citet{zhang2023crossformer}
    \item \textit{TSMixer}, \citet{chen2023tsmixer}
    \item \textit{FreTS}, \citet{yi2024frequency}
    \item \textit{Koopa}, \citet{liu2024koopa}
    \item \textit{iTransformer}, \citet{liu2024itransformer} 
    \item \textit{TimeMixer}, \citet{wang2024timemixer}
    \item \textit{TimeXer}, \cite{wang2024timexer} 
    \item \textit{MultiPatchFormer}, \cite{naghashi2025multiscale}
\end{enumerate*}

\section{Reproducibility Statement}

Our source code and documentation are already publicly available on GitHub. The Electricity and Traffic datasets are publicly available. The private MIIMIC-III dataset can be accessed by following the steps at https://mimic.mit.edu/docs/gettingstarted/. In addition, we follow the procedures outlined in previous studies to preprocess the datasets. We provide singularity and docker definitions for the container environment. We run the experiments on a single-node Linux server with 16 GB RAM and an NVIDIA RTX GPU.  We use a Python 3.12 environment with Pytorch 2.3.1 and Cuda 11.8. The random processes are seeded to ensure reproducibility. 


\end{document}